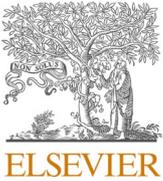
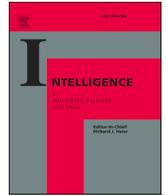

# Evidence of interrelated cognitive-like capabilities in large language models: Indications of artificial general intelligence or achievement?

David Ilić [a], Gilles E. Gignac [b],[*]

[a] *Independent Researcher, Belgrade, Serbia*
[b] *School of Psychological Science, University of Western Australia, Australia*

ABSTRACT

Large language models (LLMs) are advanced artificial intelligence (AI) systems that can perform a variety of tasks commonly found in human intelligence tests, such as defining words, performing calculations, and engaging in verbal reasoning. There are also substantial individual differences in LLM capacities. Given the consistent observation of a positive manifold and general intelligence factor in human samples, along with group-level factors (e.g., crystallised intelligence), we hypothesized that LLM test scores may also exhibit positive intercorrelations, which could potentially give rise to an artificial general ability (AGA) factor and one or more group-level factors. Based on a sample of 591 LLMs and scores from 12 tests aligned with fluid reasoning (*Gf*), domain-specific knowledge (*Gkn*), reading/writing (*Grw*), and quantitative knowledge (*Gq*), we found strong empirical evidence for a positive manifold and a general factor of ability. Additionally, we identified a combined *Gkn/Grw* group-level factor. Finally, the number of LLM parameters correlated positively with both general factor of ability and *Gkn/Grw* factor scores, although the effects showed diminishing returns. We interpreted our results to suggest that LLMs, like human cognitive abilities, may share a common underlying efficiency in processing information and solving problems, though whether LLMs manifest primarily achievement/expertise rather than intelligence remains to be determined. Finally, while models with greater numbers of parameters exhibit greater general cognitive-like abilities, akin to the connection between greater neuronal density and human general intelligence, other characteristics must also be involved.

## 1. Introduction

Gignac and Szodorai (2024, p. 4) defined artificial intelligence (AI) as "an artificial system's maximal capacity to complete a novel standardized task with veridical scoring using computational algorithms". Currently, much attention has focussed on the development and testing of large language models (LLMs), sophisticated AI systems that leverage extensive datasets and advanced neural network architectures (Zhao et al., 2023). LLMs, such as GPT, Claude, LLaMA, and Gemini, are capable of performing a wide range of tasks, including defining words, retrieving factual information, summarization, performing calculations, verbal reasoning, and creative writing.

A large number of modern LLMs have been developed since the introduction of transformer technology (Zhao et al., 2023). Like humans, LLMs exhibit substantial individual differences in capacities, as evidenced by their varied performance across a diversity of tasks (Owen, 2024). Some LLMs excel in specific domains due to specialized training data or enhanced architectural features, while others perform more robustly across a broader spectrum of tasks (Dong et al., 2023; Lin et al., 2023). This variability in performance suggests that, similar to human intelligence, there might be one or more underlying dimensions that represent LLM performance. Understanding these individual differences may prove useful for advancing our knowledge of AI capabilities and designing more effective and versatile AI systems. Additionally, scientific discoveries into the nature of AI system capabilities and behaviour could provide valuable insights into human intelligence (Gignac & Szodorai, 2024; Neubauer, 2021). Consequently, the primary purpose of this study was to examine whether LLM performance exhibits positive correlations across a range of human-like cognitive abilities. Additionally, we sought to investigate the potential emergence of one or more LLM ability dimensions, including a general ability factor.

## 1.1. General intelligence: human and artificial

Humans who tend to have higher verbal skills also tend to have higher spatial skills, better memories, and faster processing speed (Jensen, 1998). Expressed in statistical terms, the average correlation between a wide array of cognitive ability test performances tends to be approximately 0.45 to 0.50 (Detterman & Daniel, 1989; Walker et al., 2023). In general terms, the observation of consistent, positive correlations between cognitive ability test scores is known as the positive manifold (Jensen, 1998). When cognitive ability inter-correlations are submitted to data reduction procedures such as factor analysis, the largest factor tends to be a general factor that accounts for 40 to 50% of the variance in test performance (Deary et al., 2009). Known as psychometric *g* (Jensen & Weng, 1994), the phenomenon has been observed across many human cultures (Warne & Burningham, 2019), as well as several species, including orangutans (Damerius et al., 2019), dogs (Arden & Adams, 2016), and deer (Pastrana et al., 2022).

Recently, Gignac and Szodorai (2024) proposed that a general factor of ability could potentially be identified in AI systems, including LLMs, if their performance across various tasks is positively inter-correlated. It should be clarified that many computer scientists, as well the general public, commonly refer to artificial general intelligence (AGI) as a level of intelligence, specifically human-level intelligence (Amazon Web Services, 2024; Demasi et al., 2010; McLean et al., 2023; Obaid, 2023; Rayhan et al., 2023). However, because general intelligence in humans is observed across all levels of ability (Breit et al., 2022; Detterman & Daniel, 1989), Gignac and Szodorai (2024) proposed that AGI should be considered a reflection of the phenomenon of the positive manifold, rather than simply a quantitative level of cognitive ability. From this perspective, one may hypothesize that AGI is observed across all levels of AI system performance. Furthermore, by conceptualising and modeling AGI in a manner consistent with general intelligence in humans, it is possible to quantify levels of AGI across AI systems, as well as investigate predictors and outcomes of AGI.

Before proceeding, we note that Gignac and Szodorai (2024) contended that current LLMs may exhibit what they term as 'artificial achievement' (AA) rather than true AI, as these models may not fulfill all the criteria for genuine intelligence. Acknowledging that this debate is ongoing, we opted to use the broader term 'artificial general ability' (AGA) instead of the more specific term 'artificial general intelligence' (AGI) throughout this paper. We address this issue further in our discussion section.

## 1.2. Group-level factors of intelligence

Beyond the general factor of human intelligence, a reliable amount of the variance in cognitive ability tests scores is accounted for by a variety of smaller, group-level of factors, also known as stratum II dimensions (the general factor is known as a stratum III dimension; Carroll, 2003). According to the most recent Cattell-Horn-Carroll (CHC) model of intelligence, there are approximately 16 stratum II dimensions of cognitive ability (Schneider & McGrew, 2018). Of the 16 strata II dimensions, perhaps seven are relatively more dominant and commonly investigated: fluid reasoning (*Gf*), comprehension-knowledge (*Gc*), short-term memory (*Gsm*), visual processing (*Gv*), quantitative knowledge (*Gq*), cognitive processing speed (*Gs*), and reading and writing (*Grw*). We briefly review each of these seven dimensions and propose that several commonly used LLM system benchmark tests may align with some of these categories.

Fluid reasoning (*Gf*) is the ability to use various reasoning methods (e.g., inductive, deductive, analogical, etc.) to solve unfamiliar or novel problems (Kyllonen & Kell, 2017). Tests of *Gf* can be developed based on spatial, verbal, and numerical content. Raven's progressive matrices is a well-established spatial/figural test of fluid reasoning (Raven, 2000). As LLMs are based exclusively upon textual data, they may not be expected to solve spatial reasoning tasks. However, some are considered capable of solving certain types of verbal reasoning problems (Orrù et al., 2023).

Comprehension-knowledge (*Gc*) is the ability to understand and communicate culturally significant knowledge (Schneider & McGrew, 2018). A well-established measure of *Gc* is Vocabulary from the Wechsler scales (Wechsler, 2008a), as well as Similarities which measures the capacity to identify relationships between concepts. Though typically conceived as verbal tests, spatial tests of *Gc* are conceivable. For example, a geography test where participants are asked to identify nations based on their shapes may be considered a test of spatial comprehension-knowledge (e.g., Hagge, 2023). A less investigated stratum II dimension that is conceptually highly similar to *Gc* is domain-specific knowledge (*Gkn*), which represents specialized knowledge and skills in particular areas, such as knowledge of a specific academic subject (e.g., European history) or technical expertise (e.g., computer programming). Many LLM benchmark tests are would likely be classified as measures of *Gkn*, as we detail further below.

Visual processing intelligence (*Gv*) represents the cognitive ability to perceive, analyze, synthesize, manipulate, and think with visual patterns, including the capacity to understand spatial relationships. An example *Gv* test is mental rotation (Vandenberg & Kuse, 1978). As with spatial *Gf* tasks, LLMs may not be expected to solve *Gv* problems, as they are based purely on spatial content.

Quantitative knowledge (*Gq*) represents "declarative and procedural knowledge related to mathematics" (Schneider & McGrew, 2018, p. 123), which includes knowledge of symbols (e.g., $\leq$, $\infty$, $\neq$, etc.), operations (e.g., addition, multiplication, etc.) and computational procedures (e.g., long division). An example *Gq* test is the arithmetic subtest from the Multidimensional Aptitude Test II (Jackson, 2003). Though many LLMs are known to struggle with calculations (Urrutia & Araya, 2024), some can still perform reasonably well in tasks that require the recognition and manipulation of mathematical symbols and operations (Xu et al., 2024).

Short-term memory (*Gsm*) is the capacity to perceive and temporarily hold a restricted amount of information from one's present circumstances in active conscious awareness (i.e., events that transpired within the last roughly one minute). A commonly used test of short-term memory is digit span, where the participant is asked to immediately recall a sequence of randomly presented numbers in the correct order (Bowden et al., 2013). To date, there are no consistently used tests of memory span for LLMs, though LLM context windows could possibly serve as a rough proxy for short-term memory capacity (Gignac & Szodorai, 2024; Kuratov et al., 2024).

Processing speed (*Gs*) represents the capacity to execute relatively simple or well-practiced basic cognitive operations swiftly and smoothly, particularly when a high level of focused mental resources and concentration is necessitated. In humans, a well-established method to measuring processing speed is inspection time, which is a psychophysical procedure that determines the minimum exposure duration required for a person to reliably make a simple discrimination between two visual stimuli (Jensen, 2006; Nettelbeck & Lally, 1976). To date, processing speed has not been considered a dimension on which LLMs are compared, though, theoretically, it could be potentially measured (e. g., time taken by an LLM to parse and analyze large volumes of text data).

Similar to the psychometric tests used to measure the intelligence of humans, the capacities of LLMs are measured with benchmark tests, also known as datasets (Welty et al., 2019). At least conceptually, many benchmark tests can be classified with one or more CHC stratum II dimensions. For instance, Hellaswag (Zellers et al., 2019), an LLM benchmark test that involves questions related to commonsense reasoning, requiring the prediction of the most likely scenario continuation, may be regarded as a measure of *Gf*. Additionally, Winogrande (Sakaguchi et al., 2021) focuses mainly on reading comprehension test items and can be categorized as a measure of *Grw*. Finally, the HumanEval test is comprised of programming challenges and may be considered to be a measure of *Gkn*.

Over the years, numerous tests for LLMs have been developed, and many LLMs have been evaluated using benchmark tests, including the three tests noted above. Furthermore, the test results from thousands of LLMs are publicly available on the Hugging Face repository (Hugging Face, 2024). This accessibility allows for the investigation of whether LLM test performances are positively inter-correlated, potentially yielding a positive manifold. Such a positive manifold could indicate the presence of an artificial general ability factor, as well as one or more group-level factors similar to the stratum II abilities recognized within the CHC model of intelligence.

The empirical verification of a general ability factor (AGI or AGA) may be considered important, as it would provide evidence that current LLMs may possess general capabilities that extend beyond narrow task specialization (Lin et al., 2023). Additionally, the identification of a reliable general ability factor would justify the calculation and interpretation of global AI (or AA) performance scores psychometrically, which would facilitate overall performance comparisons between LLMs. Finally, by drawing parallels between the structures of human and artificial intelligence, novel insights into the nature of human intelligence may be eventually achieved (Gignac & Szodorai, 2024; Neubauer, 2021).

*1.3. Number of parameters and artificial general ability*

LLMs are fundamentally based on neural networks (Goldberg, 2016). A neural network is a computational model inspired by the way biological neural networks in the human brain process information (Jeon & Kim, 2023). It consists of layers of interconnected nodes (or neurons), where each connection has an associated weight. These weights are adjusted during the training process to minimize prediction errors and improve the model's performance. In the context of LLMs, the number of parameters refers to the total count of learnable weights and biases in the model, with each weight and bias representing a parameter that can be tuned to optimize the model's predictions (Zhao et al., 2023). For instance, RoBERTa-large is based on 355 million parameters, GPT-2-x1 on 1.56 billion parameters, and Mistral 7B on 7.3 billion parameters (Kazi & Elmahdy, 2023).

In theory, the number of parameters that underpin LLMs may be expected to positively predict the performance of LLMs. This is because a greater number of parameters allows the model to capture more complex patterns and nuances in the data, leading to a higher capacity for learning intricate relationships (Hu et al., 2021). Moreover, the greater capacity should allow the model to generalize better from training data and produce more accurate outputs across diverse tasks, potentially boosting overall performance. As a parallel, in biological organisms, including humans, there is evidence that greater neuronal density and connectivity in specific brain regions are associated with higher cognitive abilities (e.g., Dicke & Roth, 2016; Goriounova et al., 2018).

To date, a small number of studies have examined the association between model parameter size and LLM performance, though within only one model. For example, Anil et al. (2023) examined the performance of Palm 2 on 27 benchmark tests, while varying the number of model parameters experimentally, i.e., 3.86B, 7.05B, 9.50B and 16.1B parameters. The found that performance increased with larger numbers of parameters, though not linearly across all tests. From a differential psychology perspective, it would be useful to estimate the association between number of parameters and performance in a large and diverse sample of LLMs. Based on the above, we predicted a positive correlation between the number of LLM parameters and LLM test performance.

*1.4. Summary and purpose*

Many LLMs have been tested on tasks similar to those found in psychometric intelligence test batteries, and there is evidence that there are substantial individual differences in LLM capabilities. Furthermore, many LLM benchmark tests can be theoretically classified under specific stratum II dimensions within the CHC model of intelligence (Schneider & McGrew, 2018). Consequently, we investigated potential correlations between LLM system performance across diverse benchmark tests, expecting a positive manifold. We also hypothesized the existence of an artificial general factor of ability and explored the possibility of group-level factors aligning with stratum II dimensions recognized by the CHC model. Finally, in the event that artificial ability factors were observed, we hypothesized that there would be a positive association between number of LLM parameters and LLM test performance (e.g., general ability factor scores).

## 2. Method

*2.1. Sample*

The data for this study were obtained on March 8th, 2024, from the Hugging Face website, a well-known repository for LLM benchmarks. The initial sample consisted of 3862 models. However, many of the models were arguably not sufficiently distinct to be considered unique cases from an individual differences investigation perspective. Defining what constitutes a distinct model in this context is inherently complex, with no universally accepted criteria. We suggest that models may be described on a uniqueness continuum across several categories, from most unique (entirely new architectures and trained on data from scratch), moderately unique (large-scale pre-trained models, merged models with significant training adjustments, and specialized models with unique fine-tuning), less unique (fine-tuned models and merged models with minimal adjustments), and least unique (parameter-adjusted models and replicas or clones). While this continuum provides a general framework, it is important to note that the boundaries between categories can be fluid, and factors such as innovative training techniques or specialized applications can influence a model's perceived uniqueness. In our investigation, we employed three strategies to curate a sample to help reduce the impact of less distinct models on our analyses.

First, we employed Density-Based Spatial Clustering of Applications with Noise (DBSCAN; Ester et al., 1996) to remove essentially redundant models. Specifically, we used an epsilon (eps) value of 0.33 and a minimum samples parameter of 2. We arrived at these values through experimentation by collecting groups of models that were slight variations of one another and observing how they responded to changes in the DBSCAN parameters. This process reduced our sample size from 3862 to 2680 models. Next, three months after the initial data were downloaded from the repository, we found that 264 models were no longer on the site, suggesting the models were either deprecated or removed by their creators for various reasons. Finally, two models in our database possessed essentially the same name, differing only in capitalization: 'Vmware/open-llama-7b-v2-open-instruct' versus 'VMware/open-llama-7b-v2-open-instruct.' We used the most recently uploaded data for our investigation (i.e., Vmware/open-llama-7b-v2-open-instruct), resulting in a final sample size of 2415 models. Well known LLMs included those associated with Mistral, LLaMA, GPT, and Claude, as well models from less well-known sources such as ConvexAI, Lorinma, and CalderaAI, for example. We used model 'submitted time' as an approximate indicator of model age, which yielded a median of 147 days (IQR: 99 days).

For the second subsample of LLMs, we subjectively evaluated the degree of uniqueness among the 2415 models obtained in the first approach based on their names and the number of associated parameters. Information such as model architecture, fine-tuning methods, version numbers, merging or combination indicators, and specialized applications was considered, for example. Based on such an evaluation, we excluded an additional 816 models from 2415, which yielded a subsample size of $n = 1599$.

For the third and final approach, we quantified the similarity between model names among the 2415 models obtained in the first approach using a Levenshtein distance metric, which measures the number of single-character edits required to transform one name into another (Beernaerts et al., 2019; Levenshtein, 1966). Through initial trials, we determined that a stringent similarity threshold of 20 was necessary to ensure meaningful differentiation between models. Thus, any two models with a name similarity within this distance and have the same number of parameters were considered too similar. The deduplication process was implemented in Python, resulting in a final subsample size of 591 usable models. Because the results derived from all three samples were highly similar, we report only the results derived from the subsample of 591 models in the main manuscript (the results associated with the $n = 1599$ and $n = 2415$ subsamples are reported in the supplementary document).

*2.2. Measures*

A total of 12 tests were selected to represent, at least theoretically, four different group-level factors of ability: fluid reasoning (*Gf*); quantitative knowledge (*Gq*); reading/writing (*Grw*) and domain-specific knowledge (*Gkn*). Several tests were selected from the Massive Multitask Language Understanding (MMLU) test battery (Hendrycks et al., 2020). All items associated with the MMLU subtests can be viewed at https://huggingface.co/datasets/cais/mmlu.

*Gf* was measured by three tests. First, Hellaswag is a 10,042-item test designed to evaluate an LLM's abilities in natural language understanding and commonsense reasoning (Zellers et al., 2019). Test items consist of various contexts that require the identification of the most plausible continuation out of four provided options. For example, "A woman is outside with a bucket and a dog. The dog is running around trying to avoid a bath. She… a) rinses the bucket off with soap and blow dry the dog's head; b) uses a hose to keep it from getting soapy; c) gets the dog wet, then runs away again (correct); d) gets into a bathtub with the dog.

The two remaining *Gf* tests were considered measures of quantitative reasoning, a stratum I dimension of *Gf* (Schneider & McGrew, 2018). The Grade School Math 8 K (GSM8K) aims to test an LLM's ability to perform multi-step arithmetic operations and mathematical reasoning presented in a natural language context (Cobbe et al., 2021). Out of the 8500 items, we selected 1319 that best reflect quantitative reasoning. An example question is: "The ratio of boys to girls in a family is 5:7. The total number of children in the family is 180. If the boys are given $3900 to share, how much money does each boy receive?" (Answer: 52). The response format is open-ended.

The items for the third *Gf* test, specifically the Elementary Mathematics subtest within the MMLU benchmark battery (Hendrycks et al., 2020), were selected for their ability to represent quantitative reasoning. For instance, one of the items is as follows: "The rate at which a purification process can remove contaminants from a tank of water is proportional to the amount of contaminant remaining. If 20% of the contaminant can be removed during the first minute of the process and 98% must be removed to make the water safe, approximately how long will the decontamination process take? (Response alternatives: a) 2 min, b) 5 min, c) 18 min [correct], and d) 20 min.

*Gkn* was measured by a combination of three composites scores, which were based on items derived from the (MMLU) test battery (Hendrycks et al., 2020). All items associated with the MMLU subtests can be viewed at https://huggingface.co/datasets/cais/mmlu. Composite score one was based on the International Law, Business Ethics, Philosophy test items. Composite score two was based on the Medical Genetics, Clinical Knowledge, Human Aging, and Human Sexuality test items. Finally, composite score three was based on the Global Facts, Computer Security, Marketing, and Miscellaneous test items.

*Gq* was also measured by several tests within the MMLU test battery. One composite was based on the High School Statistics, Abstract Algebra, and Econometrics subtests (measuring Mathematical Knowledge; KM). A second composite was based on selected items from the

**Table 1**
Selected LLM Benchmark Tests and the Corresponding CHC Model Categorisations.

| Stratum II Ability | Stratum I Ability | Tests | Model Indicator Name | Items | α |
|---|---|---|---|---|---|
| *Gf* | Quantitative Reasoning | Elementary Mathematics | RQ | 51 | 0.74 |
| *Gf* | Quantitative Reasoning | GSM8K | GSM8K | 1319 | 0.99 |
| *Gf* | General Sequential Reasoning | Hellaswag | Hellaswag | 10,042 | 0.99 |
| *Gkn* | Law & Ethics Knowledge | International Law, Business Ethics, Philosophy | Ethics | 532 | 0.97 |
| *Gkn* | Health Science Knowledge | Medical Genetics, Clinical Knowledge, Human Aging, Human Sexuality | Health | 719 | 0.97 |
| *Gkn* | Miscellaneous Knowledge | Global Facts, Computer Security, Marketing, Miscellaneous | Misc. | 1217 | 0.87 |
| *Grw* | Reading Comprehension | HS European History | Euro. Hist | 44 | 0.96 |
| *Grw* | Reading Comprehension | HS US History | US Hist. | 48 | 0.97 |
| *Grw* | Reading Comprehension | Winogrande | Winogrande | 1267 | 0.99 |
| *Gq* | Mathematical Knowledge | High School Statistics, Abstract Algebra, Econometrics | KM | 53 | 0.93 |
| *Gq* | Mathematical Accomplishment | Elementary Mathematics | A3.E | 41 | 0.87 |
| *Gq* | Mathematical Accomplishment | Elementary Mathematics, High School Mathematics | A3.HS | 29 | 0.72 |

*Note. Gf* = fluid reasoning; *Gkn* = domain-specific knowledge; *Grw* = reading/writing; *Gq* = quantitative knowledge; α = internal consistency reliability.

Elementary Mathematics subtest (measuring Mathematical Accomplishment; A3). Finally, the third composite based on a combination of selected Elementary Mathematics items and High School Mathematics subtest items (measuring Mathematical Accomplishment; A3).

*Grw* was measured by three tests, two of which were from the MMLU: the High School European History test and the High School US History test (Hendrycks et al., 2020). For both tests, the items require carefully reading and comprehending intricate written passages, which often involve nuanced language, contextual references, and philosophical perspectives from different historical eras. Each question is multiple-choice in nature with four response alternatives. The third test, Winogrande, was classified as a measure of *Grw*, as Winogrande problems require comprehending written sentences/passages and resolving ambiguity through contextual understanding (Sakaguchi et al., 2021). Specifically, for each test item, models must choose between two options to complete a sentence, based on understanding subtle linguistic and contextual cues. For example: 'John moved the couch from the garage to the backyard to create space. The ___ is small.' Choices: a) garage (correct); b) backyard.

All tests were scored such that they represented percentage of questions answered correctly. For a list of the tests, their theorised CHC

model classifications, number of items, and test score reliabilities, see Table 1.[1]

*2.3. Data analysis*

All statistical analyses were performed using R (Version 4.2.2) and RStudio (Version 2022.09.1). To evaluate the relationships between key observed variables, Pearson correlation coefficients were calculated. We used the ggplot2 package (Wickham, 2016) to create a scatter matrix, which provided visual insights into the nature and strength of these associations. LOESS regression lines, computed with R's tricubic weighted function (Cleveland & Devlin, 1988), were included in the scatter matrix for a flexible fit, emphasizing data points based on proximity. The default setting window size of 0.75 was specified, implying that 75% of the data points in the neighbourhood of each point were used for local regression. Finally, 95% confidence intervals were included to show estimation uncertainty.

Latent variable modeling was performed using the lavaan package (Rosseel, 2012). Model fit was assessed using RMSEA and SRMR (with values ≤0.08 indicating acceptable fit), as well as CFI and TLI (with values ≥0.950 indicating acceptable fit). Given that the data were continuously scored, maximum likelihood estimation was utilized. Standardized effect 95% confidence intervals were generated through bootstrapping (5000 resamples) using a combination of the manymome (Cheung & Cheung, 2023) and semhelpinghands packages (Cheung & Cheung, 2023).

We evaluated a theoretical model comprising a second-order general factor and four first-order factors (*Gf*, *Gkn*, *Grw*, *Gq*), aligning with the categorizations and tests outlined in Table 1. To ensure proper identification and scaling, the variance of the general factor was fixed at 1, and one loading for each of the first-order factors was also constrained to 1. The data, scripts, and results are available on the OSF: https://osf.io/792ug/

## 3. Results

*3.1. Descriptive statistics and inter-subtest pearson correlations*

As can be seen in Table 2, LLM mean performance (percentage of questions answered correctly) across the 12 tests ranged from 18.09 to 74.27, with a mean of 48.62, suggesting the tests were, on average, moderately difficult. Furthermore, there were substantial individual differences in performance, as the mean test performance standard deviation was 20.02 (range: 10.99 to 31.48). The test performance distributions tended to be negative, though only substantially so for the Hellaswag test (skewness = −1.04). See Tables S1.1 and S1.2 in the supplementary document for the descriptives based on the other two subsamples.

The inter-correlations between the 12 test performance scores were all positive with a mean correlation of 0.73 (range: 0.35 to 0.99; see Table 2; see Fig. S1.1 for scatter matrix), supporting the hypothesis of a positive manifold. The *Gkn* inter-correlations were exceptionally large, ranging between 0.98 and 0.99.

*3.2. Confirmatory factor analysis*

The Kaiser-Meyer-Olkin (KMO) index analysis yielded a value of 0.93, suggesting the correlation matrix was appropriate for data reduction (Kaiser & Rice, 1974). Our hypothesized higher-order confirmatory factor analytic model did not produce an interpretable solution, as evidenced by statistically significant negative residual variances for two of the first-order factors. Consequently, we explored the structure of the data by conducting unrestricted factor analyses. First, we conducted a parallel analysis to determine the number of dimensions to extract. The parallel analysis suggested the extraction of one dimension (see Table S2.1). However, a maximum likelihood estimation factor analysis with the extraction of one factor failed to yield good model close-fit (e.g., TLI = 0.814).[2] A subsequent unrestricted factor analysis with the extraction of two factors remained unacceptably well-fitting (e.g., TLI = 0.870). Furthermore, the Hellaswag test was associated with a loading of 1.07 on the first factor, suggesting the factor solution was not stable. Moreover, the correlation between the two factors was quite large at 0.78, suggesting the presence of a substantial general factor.

Based on the above unrestricted factor analytic findings, we tested a supplementary restricted factor analytic bifactor model with a first-order general factor defined by all 12 tests. Additionally, we specified a nested, first-order *Gkn/Grw* factor. Due to the very high correlations between the Winogrande and Hellaswag tests ($r = 0.95$) and the European History and US History tests ($r = 0.97$), we also specified correlations between these respective test residuals. When we tested the model, the modification indices suggested the addition of one more correlated residual, i.e., between the A3.E and A3.HS tests. The bifactor model with the three correlated residuals yielded acceptable model close-fit, $\chi^2(44) = 250.42$, $p < .001$, CFI = 0.984, TLI = 0.977, RMSEA = 0.089, SRMR = 0.025.[3] All of the loadings were positive and statistically significant ($p < .001$; see Fig. 1). The mean general factor loading was large at 0.81 (range: 64 to 0.96; see Table S3.1 for all loadings and 95% CIs). The indicator with the largest general factor loading was associated with the KM mathematics composite ($\lambda = 0.96$). The percentage of variance accounted for by the general factor was estimated at 65.6%.[4] Furthermore, the general factor omega hierarchical was 0.94. Thus, there was evidence for a very strong general factor and, thus, highly reliable LLM general ability composite scores. Though the nested *Gkn/Grw* nested factor was notably weaker than the general factor, it was nonetheless defined by consistently positive and statistically significant factor loadings (mean $\lambda = 0.50$).

*3.3. Associations with number of parameters*

Next, we investigated the associations between the two LLM ability dimensions, artificial general ability (AGA) and *Gkn/Grw*, and the number of LLM parameters. Number of parameters data were available for 579 models. The LLMs had an average number of parameters equal to 14.86 billion. However, a histogram revealed two outliers with values of 238.09 and 180.00 billion parameters. These outliers were excluded from the correlation analyses. After removing the outliers ($n = 577$), the number of parameters had a mean of 14.19 billion ($SD = 18.87$; Median = 7.24 billion). The distribution remained skewed (skewness = 2.53). Therefore, we estimated the associations using both Pearson and Spearman correlations. Additionally, the 95% confidence intervals for the correlations were calculated using 5000 bootstrapped resamples.

The Pearson correlation between number of parameters and general

---

[1] We conducted item-total correlation analyses, prior to the testing of hypotheses. For some tests, we omitted items, as they failed to yield positive item-total correlations (see supplementary file 2). Reliabilities were estimated using item-level analyses for all tests, except for the *Gkn* composite scores, which were derived from three subtests each. For these composite scores, the three respective subtests were used as inputs in the reliability analysis.

[2] The parallel analysis procedure fails to identify the correct number of factors to extract in ≈ 2 5% of cases (Crawford et al., 2010).

[3] Despite the RMSEA being above 0.080, we considered the model to be acceptably well-fitting because the SRMR was quite low at 0.02 5. This decision is supported by simulation research indicating that RMSEA (but not SRMR) can be overly sensitive to unmodeled correlated residuals (Montoya & Edwards, 2021).

[4] We calculated the percentage of variance accounted for by summing the squared *g* loadings and dividing that sum by the total number of indicators in the model (7.87 / 12 = 6 5.6).

Table 2
Benchmark LLM test performance descriptive statistics and Pearson inter-correlations.

| Test | II | 1. | 2. | 3. | 4. | 5. | 6. | 7. | 8. | 9. | 10. | 11. | M | SD | Mdn | skew | kurtosis |
|---|---|---|---|---|---|---|---|---|---|---|---|---|---|---|---|---|---|
| 1. Hellaswag | Gf | 1.0 | | | | | | | | | | | 72.18 | 21.01 | 81.96 | −1.04 | −0.27 |
| 2. RQ | Gf | 0.45 | 1.0 | | | | | | | | | | 32.83 | 10.99 | 31.37 | 0.37 | 0.10 |
| 3. GSM8K | Gf | 0.58 | 0.62 | 1.0 | | | | | | | | | 18.09 | 21.41 | 8.61 | 1.05 | −0.21 |
| 4. KM | Gq | 0.61 | 0.73 | 0.80 | 1.0 | | | | | | | | 43.35 | 20.13 | 37.74 | 0.52 | −0.82 |
| 5. A3.E | Gq | 0.37 | 0.35 | 0.62 | 0.63 | 1.0 | | | | | | | 31.45 | 17.24 | 29.27 | 0.44 | −0.85 |
| 6. A3.HS | Gq | 0.54 | 0.56 | 0.67 | 0.73 | 0.65 | 1.0 | | | | | | 33.19 | 15.69 | 33.33 | 0.70 | 0.42 |
| 7. European History | Grw | 0.79 | 0.62 | 0.68 | 0.80 | 0.51 | 0.64 | 1.0 | | | | | 60.70 | 29.11 | 70.45 | −0.28 | −1.61 |
| 8. US History | Grw | 0.82 | 0.60 | 0.67 | 0.78 | 0.50 | 0.64 | 0.97 | 1.0 | | | | 65.34 | 31.48 | 79.17 | −0.32 | −1.64 |
| 9. Winogrande | Grw | 0.95 | 0.56 | 0.69 | 0.73 | 0.47 | 0.60 | 0.86 | 0.89 | 1.0 | | | 74.27 | 13.82 | 79.09 | −0.49 | −1.10 |
| 10. Ethics | Gkn | 0.85 | 0.65 | 0.73 | 0.84 | 0.55 | 0.70 | 0.94 | 0.95 | 0.91 | 1.0 | | 51.07 | 19.93 | 57.03 | −0.20 | −1.47 |
| 11. Health | Gkn | 0.83 | 0.68 | 0.75 | 0.86 | 0.59 | 0.71 | 0.94 | 0.95 | 0.90 | 0.98 | 1.0 | 50.05 | 19.59 | 53.98 | −0.11 | −1.50 |
| 12. Miscellaneous | Gkn | 0.86 | 0.63 | 0.71 | 0.81 | 0.55 | 0.69 | 0.95 | 0.96 | 0.91 | 0.99 | 0.98 | 50.94 | 19.80 | 58.60 | −0.32 | −1.54 |

Note. N = 591; II = theorised stratum II dimension; *Gf* = fluid reasoning; *Gq* = Quantitative Knowledge; *Grw* = reading/writing; *Gkn* = domain specific knowledge; all correlations statistically significant, *p* < .001.

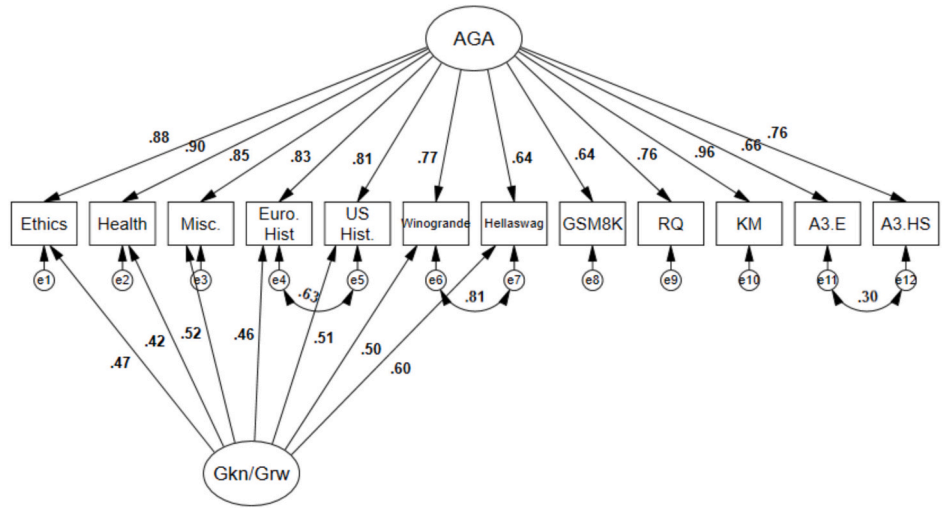

Fig. 1. Latent Variable model with completely standardized coefficients.
Note. N = 591; AGA = artificial general ability; *Gkn* = domain-specific knowledge; *Grw* = reading/writing; all coefficient statistically significant, *p* < .001; see Table 1 for model indicator descriptions; see Table S2 for 95% CIs.

ability factor scores was *r* = 0.54, 95% CI: [0.48, 0.59]. The corresponding Spearman correlation was *r* = 0.62, 95% CI: [0.60, 0.67]. A nonlinear model (cubic spline) fit the relationship between parameter count and general ability scores significantly better than a linear model, $F(2, 573) = 27.93, p < .001, \Delta\eta^2 = 0.063$. Next, we examined the nature of the association in a scatter plot with a LOESS regression line of fit. As there were only 10 LLMs with more than 80 billion parameters, we restricted the scatter plot analysis to values less than 80 billion parameters. As can be seen in Fig. 2 (left side), the analysis revealed a positive association, characterized by distinct phases. Initially, there was a sharp increase in AGA factor scores as the number of parameters increased from 100 million to approximately 10 billion parameters. This was followed by a phase of stabilization between 10 and 20 billion parameters. Beyond 30 billion parameters, the AGA factor scores continued to improve gradually and steadily up to 80 billion parameters. Overall, the trend suggests that increasing the number of parameters generally enhances general large language model capacity, with the most significant gains observed at lower parameter counts and a more gradual improvement at higher counts.

The Pearson correlation between *Gkn/Grw* and number of parameters was *r* = 0.11, 95% [CI: 0.05, 0.18]. The corresponding Spearman correlation was *r* = 0.42, 95% CI: [0.34, 0.49]. A nonlinear model (cubic spline) fit the relationship between parameter count and *Gkn/Grw* scores significantly better than a linear model, $F(2, 573) = 87.13, p < .001, \Delta\eta^2 = 0.230$. As can be seen in Fig. 2 (right side), somewhat similar to the general ability factor, the analysis revealed a positive association, characterized by distinct phases. Initially, there was a sharp increase in *Gkn/Grw* factor scores as the number of parameters increased from 0 to approximately 10 to 15 billion. This was followed by a phase of fluctuation and slight decline between 10 and 45 billion parameters, indicating variability and a non-linear relationship. Beyond 45 billion parameters, there was essentially no association between the two variables. Overall, the trend suggests that increasing the number of parameters enhances *Gkn/Grw* ability, though only up to 15 to 20 billion parameters. See Table S2.4 for a summary of the key results across all three subsamples.

## 4. Discussion

We found evidence for a LLM test performance positive manifold, as well as an artificial general ability (AGA) factor. Beyond the general factor, instead of four separate group-level factors, we found evidence for only one combined *Grw/Gkn* group-level factor. A mathematics test composite yielded the largest AGA loading. Finally, number of model parameters was found to associate positively with both AGA and the *Grw/Gkn* group-level factor. We discuss each of these key results next.

### 4.1. Positive manifold and AGA

The strength of the LLM positive manifold was remarkable, with a

D. Ilić and G.E. GignacIntelligence 106 (2024) 101858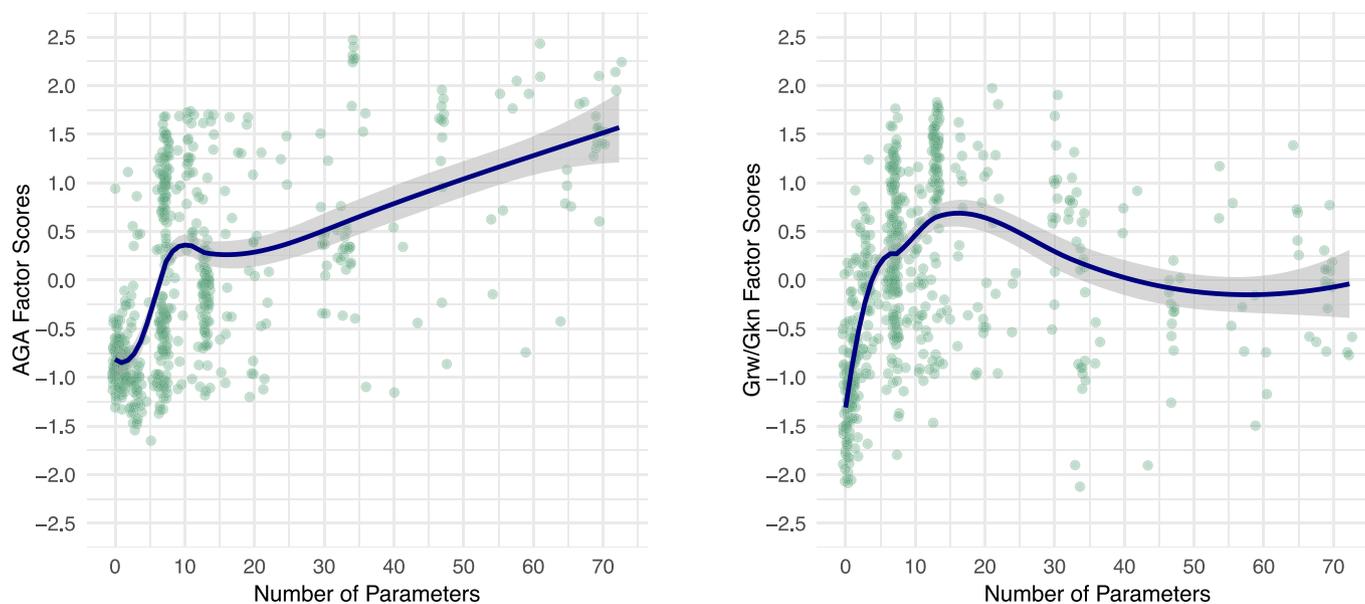

**Fig. 2.** Scatter plots depicting the association between number of parameters (billions) and artificial general ability factor scores (left side), and number of parameters and Grw/Gkn factor scores (right side).
*Note.* N = 577; lines of best fit are LOESS regression lines (tricubic weighting; span = 0.75); the jitter applied to the scatter plots was set to a width and height of 0.3 units along the respective axes; all correlations were significant, $p < .001$.

mean inter-test correlation of 0.73, significantly higher than that typically observed in humans, i.e., $r = 0.45$ to 0.50 (Detterman & Daniel, 1989; Walker et al., 2023). Correspondingly, the general factor of artificial ability accounted for 66% of the variance, whereas general factors of intelligence in human samples typically account for 40 to 50% of the variance (Deary et al., 2009). Finally, the LLM general factor yielded a coefficient of omega hierarchical of 0.94, a value higher than typically observed for the general factor of human intelligence (mid to high .80s; Dombrowski et al., 2019; Gignac, 2014a; Gignac & Watkins, 2013).

There are several possible reasons why the LLM general factor was found to be stronger than the general factor of intelligence for humans. First, the LLM test data were mostly associated with exceptional test score reliability. The median internal consistency reliability across the 12 LLM composite test scores was 0.97, whereas, as a point of comparison, the WAIS-IV technical manual (Wechsler, 2008b) reported mean subtest reliabilities of 0.88, which is probably closer to 0.75 to 0.80 in practice (see Oosterwijk et al., 2019). In contrast to LLMs, human performance is influenced by a myriad of factors including emotional and physical states, which vary widely among individuals. LLMs, however, do not experience such variability, likely leading to more reliable test performance.

Relatedly, while LLMs are trained on a vast and diverse corpus of text, the processing of this data by LLMs is uniform. Each piece of text, regardless of its origin or content, is converted into a standardized format and fed into the model through the same training algorithms and procedures (Zhao et al., 2023). This uniformity in processing ensures that the model processes the data in a consistent manner, which is in contrast to the moderate fluctuations that characterize human neurophysiology, thoughts, and behaviours over time (i.e., test-retest reliability far less than 1; Dai et al., 2019; Gnambs, 2014; Noble et al., 2019).

Finally, the median number of items per LLM test we included in the test battery was very large at 300. This is important, as the number of items in a psychometric measure is positively associated with test score reliability (Nunnally & Bernstein, 1994). Correspondingly, the internal consistencies for the LLM tests were typically very high (median 0.97), allowing for maximum possible correlations between LLM test scores (i. e., not attenuated by measurement error). By comparison, humans cannot be expected to complete very large numbers of test items without experiencing fatigue, loss of concentration, or other transient factors that can negatively impact the consistency of performance and, to some extent, the magnitude of an observed score positive manifold. We note that when restricted to the six LLM tests composed of 52 or fewer items, the mean LLM inter-test correlation was 0.63, which is smaller than the 0.73 value associated with all 12 LLM tests, but still larger than the 0.45 to 0.50 typically observed in humans (Detterman & Daniel, 1989; Walker et al., 2023). Thus, the greater breadth of coverage in some of the LLM tests composed of 100 s of items is arguably not the full explanation for why the LLMs yielded a stronger positive manifold than that typically observed in humans.

Admittedly, the range of LLM tests included in our analysis may be considered narrower compared to comprehensive measures of human intelligence. For example, none of the 12 tests assessed spatial abilities, a key dimension of cognitive ability in comprehensive batteries of human intelligence (Johnson & Bouchard Jr, 2005). Therefore, it is plausible that the percentage of variance accounted for by the LLM general factor was somewhat inflated due to the greater homogeneity in the selected LLM benchmark tests. Consequently, we refrained from labelling the LLM general factor identified in our investigation as 'artificial general intelligence,' as true AGI is expected to be demonstrated across verbal and spatial ability tests (Gignac & Szodorai, 2024; Jensen, 1998). We nonetheless note that the mean inter-subtest correlation between the verbal subtests associated with the Wechsler scales is between 0.55 and 0.60 (Wechsler, 1981; Wechsler, 2008b), which is still lower than the 0.73 mean inter-subtest correlation observed in our investigation. In simple terms, our results imply that better LLM performance on one task is substantially associated with better performance on another, much like the patterns observed in human cognitive abilities (Jensen, 1998). Thus, even considering the homogeneity of the LLM tests, the substantial shared variance across moderately distinct LLM tasks suggests the presence of general LLM capability.

### 4.2. Group-level factors

We identified only one group-level factor, not four as hypothesized. This group factor was a combination of *Gkn* and *Grw* tests. Tests designed to measure *Grw* and *Gkn* share substantial content overlap. Specifically, reading and writing skills often involve the application of general knowledge, as comprehension and production of text rely heavily on background information and vocabulary. Schneider and McGrew (2018) contended that it is beneficial to conceptualize a higher-order acquired-knowledge/expertise dimension that unites *Gc*, *Grw*, *Gkn*, and *Gq* variance. Although we did not find evidence that *Gq* tests are indicators of any factor beyond AGA, the fact that the *Grw* and *Gkn* tests formed a nested group-level factor independent of AGA is at least partially consistent with Schneider and McGrew (2018) theoretical proposition of an overarching acquired knowledge/expertise dimension.

We note that the empirical distinction between *Grw* and *Gkn* is far from clear in humans. Based on a sample of 6701 school children who completed a battery of verbal intelligence tests (*Gc*, *Grw*, and *Gkn*), Schipolowski et al. (2014) reported a latent variable correlation of 0.91 between *Grw* and *Gkn*. Additionally, based on Bryan and Mayer (2020) meta-analysis on the associations between CHC group-level factors, only one study included a *Grw* factor and none the *Gkn* dimension.[5] The correlation between *Grw* and *Gc* was reported at $r = 0.85$. Though correlations of 0.85 to 0.91 may not be entirely consistent with the notion of isomorphic constructs, it should be acknowledged that correlations of such a magnitude are so large that the apparent discriminant validity may have arisen due to method artifacts (e.g., question type) rather than substantive (construct) variance. All things considered, the observation of a combined *Grw/Gkn* dimension in our LLM data may be considered largely consistent with the human intelligence empirical literature, i.e., *Grw* and *Gkn* are highly inter-related.

We also failed to observe a *Gf* factor independent of the general factor. While some have argued that *Gf* is essentially isomorphic with *g* in humans (e.g., Gustafsson, 2001), we are more cautious about making such an interpretation with our data. Our selected measures of *Gf* were, at best, acceptable rather than good or excellent. In particular, two of our *Gf* tests focused on mathematical reasoning, and none involved figural matrices which are typically well-regarded for measuring fluid reasoning (Gignac, 2015). Consequently, further research with better measures of *Gf* is required to evaluate the possibility of a distinct *Gf* group-level factor in LLM data.

*4.3. Strongest indicator of AGA*

Much research has found that *Gf* dimensions are the strongest indicators of *g* in humans (e.g., Gignac, 2014b; Hertzog & Schaie, 1988; Kvist & Gustafsson, 2008). However, in our study, a mathematics composite that included algebra and statistics questions yielded the largest AGA loading. As previously noted, among the LLM benchmark tests available for this investigation, none were consistent with the well-regarded figural matrices tests of fluid reasoning. Consequently, it is difficult to compare our results with the broader human intelligence literature, in this context. Nonetheless, it is noteworthy that several investigations have found arithmetic to be the strongest indicator of *g* in human samples. For example, Gignac (2015) consistently found mathematics/arithmetic subtests to be the strongest indicators of *g* across bifactor analyses of three comprehensive batteries of human intelligence. Additionally, based on a bifactor analysis of the WAIS-IV normative sample, Gignac and Weiss (2015) found that the Arithmetic subtest yielded the most substantial loading onto *g*. While arithmetic may or may not be the best indicator of *g*, the literature on humans suggests that arithmetic is a strong indicator of *g*, which aligns with the results observed in our investigation with LLMs.

Based on latent variable models, several stratum II cognitive ability dimensions have been found to be unique predictors of human arithmetic ability, suggesting a relatively complex cognitive process (Floyd et al., 2003; Fung & Swanson, 2017). It is noteworthy that LLMs are known to exhibit challenges with completing arithmetic problems

(Panas et al., 2024). Possible explanations for this phenomenon include their training on diverse, noisy data which does not emphasize precision, their tendency to propagate errors in multi-step reasoning, and difficulties with symbolic manipulation (Imani et al., 2023; Qian et al., 2022; Yuan et al., 2023). Thus, LLMs capable of greater precision, error correction in multi-step problem solving, and enhanced symbolic manipulation would be expected to perform better at arithmetic, as well as a wide range of other tasks, thereby exhibiting higher levels general ability. In other words, LLMs that excel in solving arithmetic/mathematics problems, which would be expected to require more sophisticated training, model architecture, and computational algorithms, may also be superior at a variety of other verbal tasks, explaining the substantial AGA loading we observed for arithmetic.

*4.4. Association with number of parameters*

We found number of parameters to be a substantial, positive predictor of AGA, as hypothesized. Thus, our results align with the small amount of empirical research on individual LLMs that have indicated that increasing number of parameters improves LLM performance (Anil et al., 2023). We extend the literature by estimating the association based on a large sample of LLMs and a moderately diverse battery of benchmark tests. Thus, our results help support the notion that the more complex patterns captured by larger models not only facilitates better performance on specific tasks (Hu et al., 2021), but on LLM capacity in a general sense. Our finding is also consistent with research with biological organisms that has found greater cognitive abilities to be predicted by greater neuronal density and connectivity (Dicke & Roth, 2016; Goriounova et al., 2018).

While the number of model parameters accounted for approximately 25% of the variance in AGA, the relationship was clearly curvilinear. AGA factor scores generally increased with parameter count, showing the most significant gains at lower ranges (100 M to 10B), followed by somewhat weaker improvements, and essentially no association for *Gkn/Grw*, at higher parameter counts. These results are supported by Hoffmann et al. (2022), who demonstrated that for compute-optimal training, both the model size and the number of training tokens should be scaled equally. That is, increasing model parameters without also increasing the number of training tokens proportionally should not be expected to improve LLM performance uniformly. Such a perspective is somewhat analogous to neuronal density and cognitive abilities in biological organisms. While the cerebral cortex of humans and other primates has a higher neuronal density compared to other mammals, it is also the intricate patterns of connectivity and the hierarchical organization of these densely packed neurons that enable advanced cognitive functions like language, reasoning, and problem-solving (Herculano-Houzel, 2009; Roth & Dicke, 2005).

*4.5. Limitations*

First, the breadth of the tests in our 12-test battery may be regarded as limited. In particular, our 12-test battery did not include assessments from several important CHC stratum II dimensions, such as *Gv* (visual processing), *Gsm* (short-term memory), and *Gs* (processing speed). Unfortunately, none of the tests within the Hugging Face database fall into these stratum II categories. This limitation reflects the nature of the tests typically used to evaluate LLMs, which are fundamentally verbal in nature. Consequently, it is unreasonable to expect LLMs to be successful at solving visual-spatial problems. Moreover, short-term memory and processing speed have not yet been considered as dimensions for benchmarking LLMs, despite their significant roles in human intelligence (Conway et al., 2013; Jensen, 2006). It is hoped that future benchmarks will address this gap by incorporating tests that evaluate memory span and processing speed.

Additionally, as noted above, none of the included tests can be justifiably regarded as relatively pure indicators of *Gf*. While good quality *Gf* tests often involve visual items (e.g., figural matrices), there are high-quality verbal approaches to the measurement of *Gf* (Beauducel et al., 2001). For example, analogical reasoning (e.g., "Lawyer is to Client as Doctor is to ___"), inductive reasoning (e.g., determining the next letter in a sequence like "A, C, E, G, ___"), and logical reasoning (e. g., "All mammals are animals. All dogs are mammals. Therefore, all dogs are ___"). All things considered, despite the lack of breadth in our test battery, our results may be considered at least tentatively suggestive of a general factor of artificial ability, which will ideally be re-tested in future when the breadth of tests included in large scale databases is expanded.

---

[5] For reasons that are unclear, the Bryan and Mayer meta-analysis did not include Schipolowski et al.'s (2014) results.

Critics might argue that not all models in our analysis were sufficiently distinct to be considered separate cases, potentially inflating the correlations between model test performances. To address this concern, we employed three approaches to exclude insufficiently distinct cases. In our most conservative subsample, we omitted 85% of the models (3271 out of 3862), retaining only 591 for analysis. Beyond architecture (neural network layers, connections, and main components), model size, and even training data, there are many characteristics associated with a language model that can be expected to impact performance, including training duration, input representation (static versus dynamic), hyperparameters (batch sizes), objective (e.g., Masked Language Model; Next Sentence Prediction), data augmentation (artificially increase the size of the training dataset), tokenization (e.g., WordPiece; byte-level BPE), and optimization algorithms (e.g., Adam or Stochastic Gradient Descent). Thus, we believe our most conservative sample essentially represents what can be considered in practice the relatively unique models available in the first half of 2024.

As another limitation, we examined only one predictor of LLM performance. As noted, in addition to the number of model parameters, numerous other factors are expected to impact LLM performance, including the quantity and quality of training data, model architecture, tokenization, hyperparameters, and fine-tuning (Hoffmann et al., 2022). We did not have access to such information for a sufficiently large number of models to examine these factors statistically. Consequently, future research should consider examining these potential predictors as more comprehensive databases of LLM performance become available.

Finally, we acknowledge the possibility that LLMs may not exhibit true intelligence. After IBM's Watson defeated two American Jeopardy! champions in 2011, Detterman (2011) argued that this achievement did not necessarily indicate true intelligence, as that version of Watson was specifically designed to answer Jeopardy! questions and would likely perform poorly on reasoning tasks. Consequently, Detterman (2011) proposed that a more meaningful test of a computer's intelligence would involve a unique battery of IQ tests, developed by human intelligence experts. This challenge would have two levels: the first allowing data and algorithms to be supplied post hoc, similar to Watson's Jeopardy! preparation, and the second requiring only pre-programmed algorithms, forcing the computer to self-organize information as humans do. Only AI systems that answer questions on the second test would be considered to manifest true intelligence.

Although the tests utilized in the current investigation were not crafted by experts in human intelligence, there is awareness in the field that LLMs should not be specifically trained upon the benchmark tests employed to evaluate their performance (Lyu et al., 2021). Consequently, it may be posited that modern LLMs have, to some extent, met the criteria of Detterman (2011) second challenge. However, it is widely acknowledged that even more modern AI systems often struggle to consistently generalize their learned capabilities effectively (Vafa et al., 2024). Moreover, LLMs are prone to making certain types of errors that might be effortlessly avoided by humans, even those with relatively lower cognitive abilities (Tyen et al., 2023).

Considering the importance of generalizability in defining intelligence, some researchers have argued that evidence for true intelligence in AI systems remains limited (van der Maas et al., 2021). More recently, Gignac and Szodorai (2024) contended that, given the nature of LLM development and training, and the essential need for novelty in intelligence testing, there may be more evidence supporting artificial achievement/expertise than artificial intelligence. Finally, beyond the considerations of generalizability and training, some argue that true intelligence includes self-awareness and the capacity for autonomous improvement through self-evaluation (e.g., Bostrom, 2014; Mitchell, 2019; Sternberg, 2011). Correspondingly, a distinction is often made between weak AI, designed for specific tasks without true understanding or consciousness, and strong AI, which can understand, learn, and exhibit consciousness similar to human intelligence (Neubauer, 2021). Awareness and self-improvement are characteristics that were not assessed in our study of LLMs.

Thus, while our findings suggest the presence of a general factor of ability in LLMs, it is unclear whether this factor represents true artificial general intelligence or merely artificial general achievement. Regardless of the validity of Gignac and Szodorai (2024) conclusions, it is noteworthy that current LLMs exhibit a positive manifold - a phenomenon where performance on one task positively correlates with performance on others. This characteristic mirrors a fundamental property observed in human cognitive abilities (Jensen, 1998).

## 5. Conclusion

Individual differences in LLM capacities, similar to human cognitive abilities, result in a strong positive manifold. Consequently, models that perform well on one task also tend to perform well on others, suggesting the possibility of underlying general processes. Number of model parameters was found to be an appreciable, positive predictor of general and *Gkn/Grw* LLM performance. However, in the absence of other complementary model characteristics (e.g., number of tokens), the number of model parameters is likely to manifest an effect of diminishing returns. Our findings help describe the structure of artificial system capabilities and underscore the potential for further optimizing LLM performance through a balanced approach to model design. Additional work investigating LLM performance with differential psychology approaches may facilitate further advancements in both artificial and human intelligence research.

**CRediT authorship contribution statement**

**David Ilić:** Writing – review & editing, Software, Methodology, Investigation, Formal analysis, Data curation, Conceptualization. **Gilles E. Gignac:** Writing – original draft, Methodology, Investigation, Formal analysis, Conceptualization.

**Declaration of generative AI and AI-assisted technologies in the writing process**

During the preparation of this work the authors used ChatGPT, Claude, and Perplexity to help with the generation of *R* scripts, as well as to suggest superior and more concise writing. After using these tools/services, the authors reviewed and edited the content as needed and take full responsibility for the content of the publication.

**Declaration of competing interest**

The authors declare that they have no conflict of interest.

**Data availability**

Data and scripts available on the OSF link

**Appendix A. Supplementary data**

Supplementary data to this article can be found online at https://doi.org/10.1016/j.intell.2024.101858.